%% file: main.tex
\documentclass{article}

\usepackage[preprint]{neurips_2026}

\usepackage[utf8]{inputenc}
\usepackage[T1]{fontenc}
\usepackage{hyperref}
\usepackage{url}
\usepackage{booktabs}
\usepackage{multirow}
\usepackage{graphicx}
\usepackage{amsmath}
\usepackage{amssymb}
\usepackage{amsfonts}
\usepackage{nicefrac}
\usepackage{microtype}
\usepackage{xcolor}
\usepackage{cleveref}
\usepackage{wrapfig}
 \usepackage{subcaption}

\input{math_commands.tex}

\title{Attention Quantization for Tabular Foundation Models}

\author{%
  Jonas M.~K\"ubler\thanks{Correspondence: \texttt{jonas@priorlabs.ai}} \\
  Prior Labs \\
  \And
  Benjamin J\"ager \\
  Prior Labs \\
  \And
  Klemens Fl\"oge \\
  Prior Labs \\
  \AND
  Noah Hollmann \\
  Prior Labs \\
  \And
  Frank Hutter \\
  Prior Labs \\
}

\begin{document}

\maketitle

\begin{abstract}
\input{chapters/abstract}
\end{abstract}

\input{chapters/introduction}

\input{chapters/related_work}
\input{chapters/method}
\input{chapters/experiments}

\input{chapters/conclusion}

\bibliography{references}
\bibliographystyle{plainnat}

\appendix
\input{chapters/appendix}

\end{document}

%% file: math_commands.tex
\usepackage{amsmath,amsfonts,bm}

\def\eqref#1{equation~\ref{#1}}

\def\1{\bm{1}}

\DeclareMathAlphabet{\mathsfit}{\encodingdefault}{\sfdefault}{m}{sl}
\SetMathAlphabet{\mathsfit}{bold}{\encodingdefault}{\sfdefault}{bx}{n}

%% file: chapters/abstract.tex
With the recent rise and adoption of tabular foundation models, 
optimizing their inference performance becomes an emerging
field for efficiency research.
While the models are architecturally similar to transformer-based
large language models (LLMs), the size and serving patterns differ significantly.
We show that the focus should be on the attention \textit{calculation} and less on weight or KV cache quantization, which are more popular in LLMs.
We develop a quantization strategy for queries, keys, and values
to FP8 and use explicit FP8 matrix multiplication 
instructions 
to speed up the
attention calculation. We find that it is crucial to align the quantization error in the test rows with the quantization error in the training rows, as otherwise the accuracy drops drastically. Our 
Triton kernel achieves a speedup up to 1.7x over regular 
16-bit kernels, and we show that on TabPFN-v3  and TabICLv2 there is no relevant accuracy loss 
across TabArena and BeyondArena.

%% file: chapters/introduction.tex
\begin{figure*}[!h]
\centering
\begin{minipage}[t]{0.50\textwidth}
  \vspace{-20pt}
  \centering
  \includegraphics[width=\linewidth]{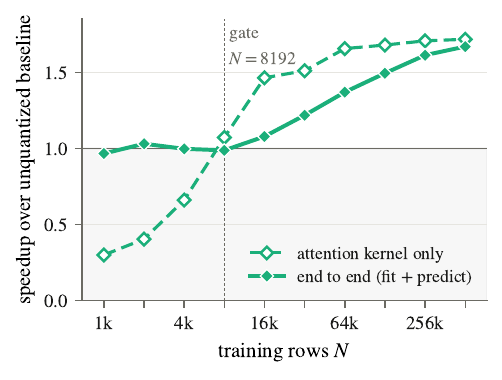}
  \vspace{-20pt}
\end{minipage}
  \vspace{4pt}
\begin{minipage}[t]{0.48\textwidth}
  \vspace{0pt}                     %
  \centering
  \small
\begin{tabular}{lcc}
  \toprule
                     & Baseline & FP8 ($>8192$)\\
  \midrule
  \multicolumn{3}{l}{Elo} \\
  \quad TabArena     & $1648.7 \pm 2.1$ & $1648.4 \pm 2.1$ \\
  \quad BeyondArena  & $1269.6 \pm 2.0$ & $1269.1 \pm 2.2$ \\
  \addlinespace
  \multicolumn{3}{l}{Benchmark wall-clock (min)} \\
  \quad TabArena     & $29.6 \pm 0.3$   & $29.4 \pm 0.3$ \\
  \quad BeyondArena  & $253.8 \pm 7.1$  & $183.7\pm 0.9$ \\
  \bottomrule
\end{tabular}
\end{minipage}
\caption{FP8 attention kernel and end-to-end speedup (left). The kernel is turned on when there are more than 8192 train rows. The accuracy impact (right) on TabArena \citep{erickson2026tabarena} and BeyondArena \citep{purucker2026beyond} is within seed noise, and we observe large wall-clock time speedups for the entire evaluation suite on BeyondArena including data preparation and scoring. 
}
\label{fig:teaser}
\end{figure*}

\section{Introduction}

Transformer-based models \citep{vaswani2017attention} are now widely adopted in large language models (LLMs) and recently
also in tabular foundation models.
Their broad adoption makes it necessary to reduce inference costs and latency. 
For LLMs, quantization \citep{frantar:2023:gptq, parkInference2024, kurtic-etal-2025-give} has been one of the most successful approaches to cut costs and latency.

For tabular foundation models, such as the TabPFN \citep{hollmann2025accurate, grinsztajn2026tabpfn} or TabICL \citep{qu2025tabicl, qu2026tabiclv2} model families, the models themselves are still fairly small, so weight quantization \citep{luo2026memoryefficienttabularfoundation} offers limited benefits. 
We instead explore quantizing the matrix multiplications in the fused attention. 
This is particularly helpful for tabular foundation models as their cost scales quadratically with the
row count of the training dataset, which can be hundreds of thousands or even millions. However, it is unknown how the predictive performance is affected.

Modern hardware accelerators like NVIDIA's Hopper and Blackwell GPUs, AWS' Trainium 3, or AMD Instinct MI325X offer instructions for FP8 matrix multiplications that have twice the throughput over their 16-bit counterparts. 
While speedups can be realized in standalone large matrix multiplications, it is not trivial to realize them in a fused attention kernel, as other operations, like exponentiation, can become the bottleneck \citep{zadouri2026flashattention4algorithmkernelpipelining}. Whether FP8 matrix multiplications can lead to speedups thus depends on the architecture and hardware specifications.

To the best of our knowledge, this is the first study of FP8 attention in the context of tabular foundation models. We use TabPFN-v3 \citep{grinsztajn2026tabpfn} as the workhorse, and our results motivate using models with FP8 attention as a plug-in post-training. First, we show that a per-head absmax-scaled quantization of queries, keys, and values in the in-context learning (ICL) layer of the transformer can be done without relevant accuracy degradation. Remarkably, this is nearly lossless only when \textit{both} the train--train and the test--train attention are quantized. 
Second, we show that on an NVIDIA RTX Pro 6000 we can achieve up to 1.7x kernel speedups through FP8. 
While our ablations are centered around one model, we also evaluate on TabICLv2 \citep{qu2026tabiclv2} and obtain similar accuracy and speed results.

%% file: chapters/method.tex
\section{Method}
\label{sec:method}
A softmax attention head maps queries
$Q \in \mathbb{R}^{S_q \times d}$, keys $K \in \mathbb{R}^{S_{kv} \times d}$, and values
$V \in \mathbb{R}^{S_{kv} \times d}$ to
\begin{equation}
  \label{eq:attention}
  \mathrm{Attn}(Q, K, V) = P V, \qquad 
  \text{with }  P = \mathrm{softmax}\!\left(\alpha\, Q K^\top\right),
\end{equation}
where the softmax is taken row-wise and $\alpha$ is a scalar.
Since the head dimension $d$ is usually small, 
materializing the full attention matrix $P$ in global memory is a bottleneck.
Therefore, fused implementations, like Flash Attention (FA) \citep{dao2022flashattention}, never materialize
$P$ in memory: they tile the sequence axis and accumulate the output in a single
pass, so the two matrix multiplications
$QK^\top$ and $PV$ dominate the runtime 
of the
layer. In this work we quantize these matrix multiplications.

In a tabular foundation model the primary attention bottleneck is the in-context
learning (ICL) stage \citep{qu2026tabiclv2,hollmann2025accurate}, which attends across rows of the dataset.
The ICL stage is not a single attention call \citep{grinsztajn2026tabpfn}. Writing $N$ for the number of
training rows and $M$ for the number of test rows, keys and values are projected
from the training rows alone, while queries are projected from all rows. The layer
therefore evaluates two independent instances of \Cref{eq:attention}: a
\emph{train--train} attention in which the $N$ training queries attend to the $N$
training keys, and a \emph{test--train} attention in which the $M$ test queries
attend to the same $N$ training keys.
Overall, since the embedding dimension of the ICL stage is independent of features, the overall FLOPs of the model forward pass grow as 
$\mathcal{O}(N^2 +MN)$. In contrast, the FLOPs associated with the weight matrix multiplications only grow linearly 
$\mathcal{O}(N+M)$, so FP8 attention targets the dominant bottleneck that determines runtime for large $N$, see \Cref{fig:noise_and_latency}.

We quantize to the \texttt{e4m3fn} FP8 format
\citep{micikevicius2022fp8}, whose range is $[-448, 448]$. To stay within the representable range, each quantized tensor $X$ is
represented as $\hat{X} \approx s_X X_8$ with $X_8$ in FP8 and an FP32
dequantization scale $s_X$.  We compute the scale dynamically from the tensor itself rather than using a static calibration. For each head $h$ we take the absolute maximum,
  \begin{equation}                                                           
    \label{eq:amax-scale}                                                    
    s_X^{(h)} = \frac{1}{448}\max_{i,j} \bigl| X^{(h)}_{ij} \bigr|,          
    \qquad                                                                   
    X_8 = \mathrm{rn}_{\texttt{fp8}}\!\left(                              
          \mathrm{clamp}\bigl( X / s_X,\, -448,\, 448 \bigr)\right),         
  \end{equation}                                                             
  where $\mathrm{rn}_{\texttt{fp8}}$ rounds to the nearest representable value. By construction the largest entry of each head
 maps to $\pm448$, so the range is used in full. 
  $P$ is the exception as it is not materialized outside of the kernel.
However, since it is the softmax output, its range is \emph{known} to be 
in $(0, 1]$, which wastes the exponent range above $1$ and pushes the small
probabilities into the subnormals. To circumvent this, we add a global
constant to the exponent before exponentiation, which lifts the row maximum to $2^8 = 256$,
below the $448$ ceiling, as is also done in FA3 \citep{shah2024flashattention}.

For the test--train attention, $K$ and $V$ are from the training rows and thus keep their scales. 
However, computing a dynamic scale on test queries \textit{couples the computation of the test rows}. 
Although this does not measurably affect accuracy, it makes predictions depend on the exact test batch.
We thus propose to use the scale computed on the train queries and reuse this for the test queries. Our experiments show that this works equally well and makes the test predictions batch invariant.

\begin{figure}[t]
    \centering
    \includegraphics[width=\linewidth]{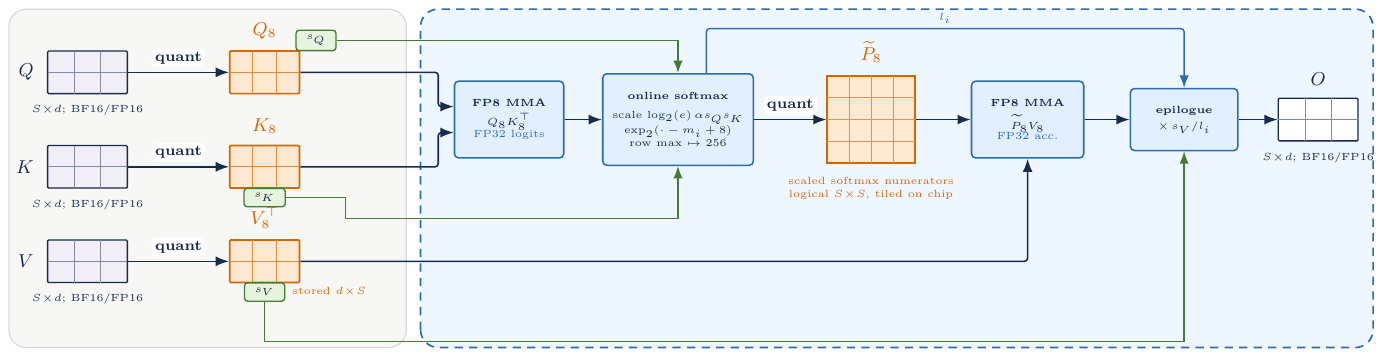}
    \caption{Quantization approach illustrated for a single head. We quantize $Q, K, V$ with per-head scaled absmax quantization, the $Q$ and $K$ scales are then applied in the kernel together with the softmax scale, whereas the $V$ scale is applied in the epilogue. The $P$ matrix is never materialized in global memory and is quantized with a fixed scale inside the kernel.}
    \label{fig:kernel_workflow}
\end{figure}
The baseline for our kernel is FA2 \citep{dao2023flashattention} called via torch scaled dot product attention (SDPA). Note that FA3 or FA4 do not apply to RTX Pro 6000. Our kernel is a Triton kernel and very closely follows the overall design of FA2 with the quantization similar to FA3 \citep{shah2024flashattention}, see \Cref{fig:kernel_workflow}. While there exist other low-precision attention kernels for other architectures \citep{zhang2025sageattention, zhang2025sageattention2, shah2024flashattention}, we note that previous low-precision kernels for Hopper and Ampere architectures had to work around an (implicit) low-precision accumulator \citep[Section 3.4]{zhang2025sageattention2}, which costs performance, but is not required for Blackwell chips we are targeting. \Cref{fig:teaser} compares the kernel speed to the FP16 baseline. The quantization overhead is amortized for $N>8192$, so our final proposal gates on that.

%% file: chapters/experiments.tex
\section{Experiments}
For our main experiments we use TabPFN-v3 \citep{grinsztajn2026tabpfn} and defer results on TabICLv2 \citep{qu2026tabiclv2}, which confirm our findings, to the appendix. While our final proposal gates on $N>8192$, \Cref{sec:TabArena_ablations} first investigates different strategies for quantization without the gate to investigate the accuracy impact of quantization independently of its latency aspects. All our experiments are run on RTX Pro 6000 Blackwell edition. For context, models on the   \href{https://huggingface.co/spaces/TabArena/leaderboard}{TabArena leaderboard} are typically separated by more than 10 Elo points.

\subsection{Quality impact on TabArena}\label{sec:TabArena_ablations}
We first test the quality impact of our quantization scheme and kernel on TabArena \citep{erickson2026tabarena}, where we run the full suite consisting of 51 datasets. We experiment with applying FP8 to train$\leftrightarrow$train only, to test$\leftrightarrow$train only, or both, where we also distinguish between using the scale from the train queries also for the test queries. 
In addition to the effect size,  we also compute a $p$-value for the null hypothesis that the model did not get worse. Following \citet{kubler2026when} we assume that post-training quantization can only make the model worse and use a \textit{one-sided} sign test. We run over three seeds of TabPFN-v3's preprocessing. 

The aggregated results are in \Cref{tab:fp8-ablation} and the full results in \Cref{tab:appendix-fp8}. We find that quantizing both attention calls minimizes the degradation, and the drops are almost negligible in practice. That said, our strict statistical treatment reveals that this tiny Elo drop likely is a small but \textit{detectable} regression when the kernel is applied without gate. Furthermore, sharing the query scale from the train rows with the test rows does not alter the results in a relevant way, but makes predictions batch invariant, which is the approach we thus choose for the final experiments.

\begin{table}[t]
\centering
\caption{%
FP8 attention ablations on TabPFN-v3 with TabArena without $N>8192$ gating. 
}
\label{tab:fp8-ablation}
\begin{tabular}{lccc}
\toprule
FP8 applied to & $\Delta$ error (\%) & $\Delta$ Elo & sign-test $p$ \\
\midrule
train$\leftrightarrow$train only & $+2.81 \pm 0.73$ & $-31.8 \pm 4.1$ & $<$0.001--0.012 \\
test$\leftrightarrow$train only & $+2.38 \pm 0.77$ & $-20.3 \pm 2.4$ & $<$0.001--0.005 \\
both, per-call query scale & $+0.01 \pm 0.03$ & $-1.0 \pm 0.4$ & 0.024--0.131 \\
\textbf{both, shared train query scale} & {\boldmath$+0.02 \pm 0.03$} & {\boldmath$-1.3 \pm 0.3$} & \textbf{0.024--0.080} \\
\bottomrule
\end{tabular}
\end{table}

\subsection{Coordination between self-attention and cross-attention is needed}

\begin{figure}[t]
\vspace{-5pt}
    \centering
    \begin{subfigure}[t]{0.49\linewidth}
      \centering
      \includegraphics[width=\linewidth]{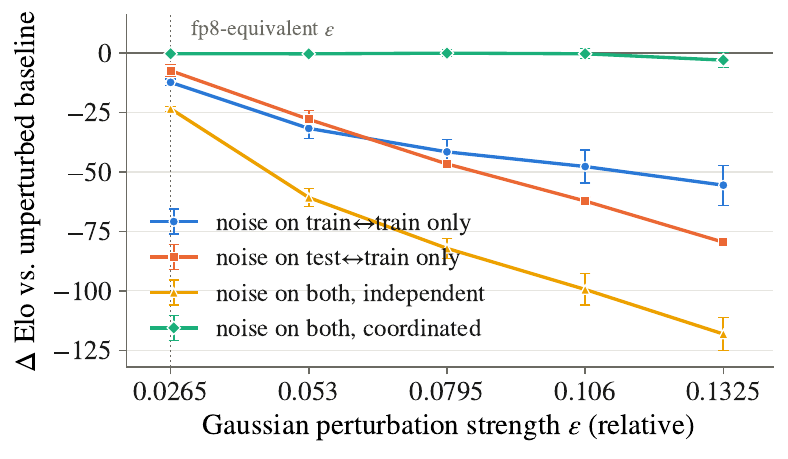}

      \label{fig:noise-sweep}
    \end{subfigure}\hfill
    \begin{subfigure}[t]{0.49\linewidth}
      \centering
      \includegraphics[width=\linewidth]{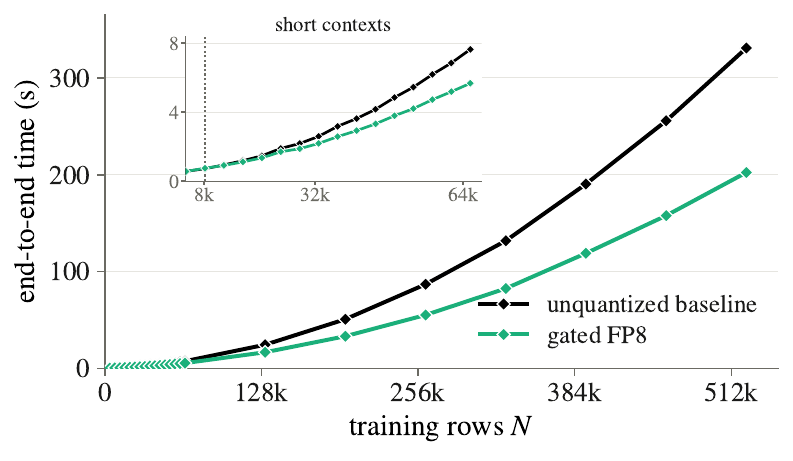}
      \label{fig:e2e-runtime}
    \end{subfigure}
    \caption{TabArena performance across three random seeds when Gaussian noise is
    added to $Q, K, V$ (left) and E2E runtime scaling.}
    \label{fig:noise_and_latency}
  \end{figure}
In \Cref{tab:fp8-ablation} we find that it is critical to apply the same perturbation to both train and test rows. This might be counterintuitive, as we apply more \textit{local} perturbation than if we just quantize one call.
This is a remarkable feature of the architecture and to investigate whether it holds beyond quantization we conduct further experiments.
We add random Gaussian noise to queries, keys, and values of the ICL layer, both in the individual calls only as well as on both calls. When applying it to both calls, we ablate independent noise as well as applying the same noise to $K$ and $V$ across attention calls (coordinated), see \Cref{app:noise}. We then evaluate over TabArena with 3 different random seeds. The results are shown in \Cref{fig:noise_and_latency}, and demonstrate that this is a model's more general ability to sustain errors on the attention inputs as long as both attention calls receive the same version of $K$ and $V$.

\subsection{Latency measurements and final configuration experiments}
We investigate the kernel latency and the resulting end-to-end latency. Quantization amortizes around 8,192 training rows. The kernel speedups reach 1.72x and the end-to-end speedups up to 1.67x. Full results are shown in \Cref{tab:speed} and \Cref{fig:teaser} visualizes the relative speedups. \Cref{fig:noise_and_latency} further shows the quadratic runtime scaling and the large absolute gains of using FP8. For completeness we also ablate a 16-bit variant of our kernel to ensure the speedups result from FP8 computation and not just better kernel design. Results are in \Cref{tab:speed} and show that our 16-bit variant closely follows the FA2 baseline.
To show the generality of our kernel, in \Cref{app:generalization} we also investigate head dimension 128 and L4 GPUs, for which our kernel provides even larger speedups.

For the final runs we activate the kernel only if there are more than 8,192 train rows (gate) and share the query scale computed from the train rows with the test queries, to make the test predictions batch invariant. With that configuration we run TabArena again and also BeyondArena \citep{purucker2026beyond}, again using three TabPFN preprocessing seeds. BeyondArena has 142 datasets out of which 56 are above the 8,192 gate. In particular it also has datasets with 1M training rows. Thus on this benchmark, while being essentially on par with the baseline, we see a substantial wall clock speedup, see \Cref{fig:teaser} and \Cref{tab:appendix-fp8}, and Appendix \ref{app:per-dataset} for a per-dataset resolution of the errors, showing that FP8 errors are smaller than preprocessing seed variations.

%% file: chapters/conclusion.tex
\section{Conclusion and Outlook}
We introduced attention quantization to tabular foundation models and showed that for datasets containing hundreds of thousands training rows, it can speed up inference end-to-end by up to 1.7x without relevant quality degradation on TabArena and BeyondArena. While faster prediction time increases usability, quantization also cuts the cost by a similar factor.
This opens a new direction for post-training optimization of tabular foundation models and we found that coordination between train and test rows is required, a concept that is not present in LLMs.

%% file: chapters/appendix.tex
\section{Further Experimental Results and Details}
\subsection{Details on the Gaussian noise experiment}
\label{app:noise}
In the TabPFN-v3 architecture, test rows attend to the first $K$,$V$ head in a grouped-query attention style. The critical insight we find is that the calls should not receive different versions of that $K$,$V$-pair.

To analyze this further, in \Cref{fig:noise_and_latency} we replaced quantization by an
explicit random perturbation, so that we can the type of perturbation as well as its size. For a tensor $X \in \mathbb{R}^{S \times H \times d}$  ($H=8$ heads) and a strength $\varepsilon > 0$ we replace $X$ by
\begin{equation}
  \label{eq:noise}
  \tilde{X}^{(h)} = X^{(h)} + \varepsilon \, \rho^{(h)} Z^{(h)},
  \qquad
  \rho^{(h)} = \Bigl( \tfrac{1}{S d} \textstyle\sum_{i,j} \bigl(X^{(h)}_{ij}\bigr)^2 \Bigr)^{1/2},
  \qquad
  Z^{(h)}_{ij} \stackrel{\text{iid}}{\sim} \mathcal{N}(0, 1),
\end{equation}
independently for every head $h$. The scale is set per head by that head's own
RMS $\rho^{(h)}$, mirroring the per-head scale $s_X^{(h)}$ of
\Cref{eq:amax-scale}. With this, heads whose entries differ by orders of magnitude are
perturbed equally in relative terms. We perturb $Q$, $K$ and $V$
individually, and every attention layer draws its own field from a seeded
generator, so a run is reproducible from the seed alone.

We sweep $\varepsilon \in \{1,2,3,4,5\} \times 0.0265$, where $0.0265$ is the error we found to usually arise from quantization. We thus swipe starting from an error similar to the quantization error. 

We then ablate the following configurations:
\begin{itemize}
  \item \textbf{train--train only:} Only the train--train call receives
    perturbed inputs, while the test--train call gets $Q$, $K$, $V$ exactly.
  \item \textbf{test--train only:} Only the test--train call receives
    perturbed inputs, while the train--train call gets $Q$, $K$, $V$ exactly.
  \item \textbf{both, independent:} Both calls receive perturbed inputs, but
    the test--train call independently perturbs its $K$ and $V$. Therefore the two
    calls see two different noisy versions of the same training keys and
    values.
  \item \textbf{both, coordinated:} Both calls receive perturbed inputs, and
    the perturbation of $K$ and $V$ is applied \emph{once, before the head
    slice}. The test--train call then consumes exactly the $K,V$ entries the
    train--train call consumed, whilst the $Q$ is still perturbed independently.
\end{itemize}

\subsection{Per-dataset behavior of the gated final configuration}
\label{app:per-dataset}

\begin{figure}[t]
  \centering
  \includegraphics[width=\linewidth]{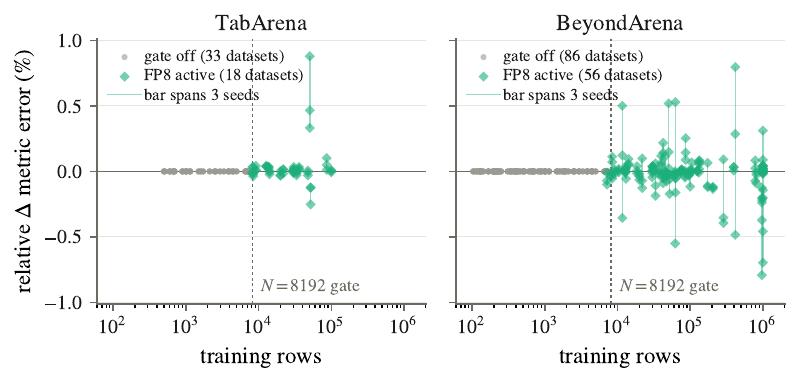}
  \caption{Per-dataset relative change in metric error of the gated FP8
  configuration against the unquantized baseline, over three preprocessing
  seeds. Grey points are datasets below the $N>8192$ gate, which run the
  unquantized path and match the baseline. Green points are
  those where FP8 is active. Each bar spans the three seeds.}
  \label{fig:per_dataset}
\end{figure}

\Cref{fig:per_dataset} shows where the aggregate numbers of
\Cref{tab:appendix-fp8} come from. We show one point per dataset and seed, for the
final configuration (gate at $N>8192$, shared train query scale) against the
unquantized baseline at the same seed.

Below the gate, FP8 is not activated and the results match exactly. Above it, FP8 is active on $18$ of $51$
TabArena datasets and $56$ of $142$ BeyondArena datasets. There is no trend with dataset size, so for example, large datasets do not degrade more, although their speedup is larger. 
The gate is applied per run rather than per dataset. Since some datasets have splits that
differ in size, it can engage on some splits and not others. We plot those datasets at their median size, which is why a marker can
be left of the threshold. 

We report relative error because the Arenas mix \texttt{roc\_auc},
\texttt{log\_loss} and \texttt{rmse}, whose absolute errors differ by orders of
magnitude. The cost is that tiny movements on datasets that are almost optimal are inflated. The largest case in TabArena is APSFailure at around 50k training rows, scored by
\texttt{roc\_auc} at a baseline error of $0.0038$--$0.0069$, i.e.\ AUC
$0.993$--$0.996$, where a relative error of $+2\%$ on a single split is an
absolute AUC change of $8\cdot 10^{-5}$. Furthermore, if we compare the error induced by TabPFN preprocessing over seeds, it is around five times larger than the within-seed FP8 deviations.

We further note that for BeyondArena, although there is a slight mean Elo drop, the FP8 model on average wins on the relative error. We do not claim though that the model got better, but attribute this to noise.

\input{tables/appendix_results}

\input{tables/speed_table}

\subsection{Experiments on TabICLv2}
In the main body of the paper we did thorough ablations on TabPFNv3. To test whether FP8 attention works also on other models, and thus does not depend on specific architecture sizes, prior, and optimization choices, we also evaluate on TabICLv2 \citep{qu2026tabiclv2}. 
While TabPFNv3 has grouped query attention for the test rows and thus there naturally are two separate attention calls, TabICLv2 uses the same multi-head attention for both train and test rows. For simplicity we thus also use only a single attention call which we quantize with per-head dynamic scales. 

For simplicity we also just use one implementation without gating on $N>8192$ as this allows us to investigate the accuracy impact as a whole, which is interesting scientifically even if there is (currently) no speedup in this regime. There is one more subtlety which is that TabICLv2 runs very small datasets in FP32. To not confound our insights about FP8 quantization with moving from 32 bits to 16 bits, we do not enable the kernel for those datasets. 
We show the accuracy results in \Cref{tab:tabicl-fp8} and in \Cref{tab:tabicl-speed} and they confirm our insights from TabPFN-v3. The attention quantization is nearly lossless, where in this case the $p$-value is not significant. And the end to end speedup gains are also similar, reaching 1.56x on 512k train rows, which illustrates that on TabICLv2 other operations take a larger share of the overall latency than on TabPFNv3.

\input{tables/tabicl_fp8}
\input{tables/tabicl_speed}

\subsection{Generalization to different head dimensions and hardware}\label{app:generalization}
In the main paper we focused on RTX Pro 6000 and on head dimension 64 as this is used by both  TabPFN-v3 and TabICLv2. To show that our kernel speedups generalize across hardware platforms and architectures, we also test our kernel on NVIDIA L4 GPUs as well as on head dimension 128. 
The results for head dimension 128 are shown in \Cref{tab:three-way-d128} and show speedups up to 1.82x which is larger than for the 64 case. The explanation is that the larger the head dimension the more the matrix multiplications dominate and other operations like application of scales or exponentiation matter less. 

For the L4 benchmarks on head dim 64 in \Cref{tab:three-way-l4-d64} we even observe speedups over torch's SDPA of up to 1.89x, though some of this can be attributed to its kernel not being perfectly optimized, as our own 16-bit variant is faster.

This brings us to limitations of our current kernel. For hardware like B200 or H100 the kernel would need to use special hardware units like the Tensor Memory Accelerator (TMA) to even reach 16 bit performance \citep{shah2024flashattention}. Since our kernel is an FA2 design, it does not use these. Furthermore, on B200 GPUs the tensor core performance has been scaled so drastically that even in 16 bit compute, the softmax can become the bottleneck \citep{zadouri2026flashattention4algorithmkernelpipelining}. In this case simply speeding up the matrix multiplications through FP8 would not result in meaningful speedups at all. 
\include{tables/three_way_d128}
\include{tables/three_way_l4}

%% file: tables/appendix_results.tex
\begin{table}[t]
\centering
\small
\caption{%
  Accuracy ablations on \textbf{TabPFN-v3}. We run all the experiments over three seeds in TabPFN, whereas the datasets are not random.
  \emph{$n$} is the number of datasets whose predictions differ from
  the baseline at all: the gated arm leaves those below $8192$ training rows untouched and hence they reproduce the baseline exactly. They are not considered in the sign test. \emph{win} is the fraction of those $n$ on which the
  arm beat the baseline, so $0.5$ is the null hypothesis. This is the quantity the
  sign test operates on. $\Delta$err is the mean per-dataset
  relative change in \texttt{metric\_error}, positive being worse.
  $p$ is the one-sided sign test $p$-value.%
}
\label{tab:appendix-fp8}
\begin{tabular}{llrrrrr}
\toprule
& seed & $n$ & win & $\Delta$err (\%) & $\Delta$Elo & sign $p$ \\
\midrule
\multicolumn{7}{l}{\textit{TabArena, all splits (51 datasets)}} \\
train$\leftrightarrow$train only & 0 & 51 & 0.22 & $+2.599$ & $-29.8$ & $<$0.001 \\
 & 1 & 51 & 0.33 & $+2.206$ & $-29.2$ & 0.012 \\
 & 2 & 51 & 0.22 & $+3.623$ & $-36.5$ & $<$0.001 \\
\addlinespace
test$\leftrightarrow$train only & 0 & 51 & 0.29 & $+2.085$ & $-21.2$ & 0.002 \\
 & 1 & 51 & 0.24 & $+1.802$ & $-17.5$ & $<$0.001 \\
 & 2 & 51 & 0.31 & $+3.254$ & $-22.1$ & 0.005 \\
\addlinespace
both, per-call query scale & 0 & 51 & 0.35 & $+0.029$ & $-1.4$ & 0.024 \\
 & 1 & 51 & 0.41 & $-0.024$ & $-1.1$ & 0.131 \\
 & 2 & 51 & 0.35 & $+0.027$ & $-0.6$ & 0.024 \\
\addlinespace
both, shared train query scale & 0 & 51 & 0.37 & $+0.034$ & $-1.6$ & 0.046 \\
 & 1 & 51 & 0.35 & $-0.013$ & $-1.2$ & 0.024 \\
 & 2 & 51 & 0.39 & $+0.027$ & $-1.1$ & 0.080 \\
\addlinespace
\textbf{both, shared scale, gated} & 0 & 18 & 0.33 & $+0.009$ & $-0.1$ & 0.119 \\
 & 1 & 18 & 0.44 & $+0.003$ & $-0.5$ & 0.407 \\
 & 2 & 18 & 0.28 & $+0.020$ & $-0.3$ & 0.048 \\
\midrule
\multicolumn{7}{l}{\textit{BeyondArena \textup{core} (142 datasets)}} \\
\textbf{both, shared scale, gated} & 0 & 56 & 0.46 & $+0.004$ & $-2.0$ & 0.344 \\
 & 1 & 56 & 0.52 & $-0.010$ & $+0.6$ & 0.656 \\
 & 2 & 56 & 0.59 & $-0.011$ & $-0.1$ & 0.930 \\
\bottomrule
\end{tabular}
\end{table}

%% file: tables/speed_table.tex
\begin{table}[t]
\centering
\caption{%
  Speed of the kernel and end-to-end on TabPFN-v3 on NVIDIA RTX PRO 6000 Blackwell Server Edition. Our baseline is always torch SDPA, which uses FlashAttention 2. But for comparison we also include a 16-bit version of our own kernel, which closely follows torch. Therefore, the speedups can be clearly attributed to usage of FP8 tensor cores.
  \emph{Kernel}: a single ICL train$\leftrightarrow$train attention call
  ($B=1$, $H=H_{kv}=8$, $D=64$, float16), where quantization of $Q$, $K$, and $V$ is included.
  \emph{End-to-end}: \texttt{predict} of a 20-feature, 1024-test-row problem.
  FP8 is gated off below $8192$ training rows in the end-to-end tests.%
}
\label{tab:speed}
\begin{tabular}{rrrrrrrr}
\toprule
& \multicolumn{4}{c}{attention kernel (ms)} & \multicolumn{3}{c}{end to end, predict (s)} \\
\cmidrule(lr){2-5}\cmidrule(lr){6-8}
$N$ & \texttt{torch} & ours 16-bit & ours FP8 & FP8 speedup & baseline & FP8 & speedup \\
\midrule
1k & 0.065 & 0.049 & 0.216 & 0.30$\times$ & 0.183 & 0.189 & 0.97$\times$ \\
2k & 0.097 & 0.069 & 0.239 & 0.40$\times$ & 0.282 & 0.274 & 1.03$\times$ \\
4k & 0.195 & 0.177 & 0.295 & 0.66$\times$ & 0.322 & 0.322 & 1.00$\times$ \\
\midrule
8k & 0.504 & 0.471 & 0.470 & 1.07$\times$ & 0.495 & 0.501 & 0.99$\times$ \\
16k & 1.781 & 1.601 & 1.216 & 1.46$\times$ & 0.966 & 0.895 & \textbf{1.08$\times$} \\
32k & 5.983 & 5.802 & 3.962 & 1.51$\times$ & 2.361 & 1.938 & \textbf{1.22$\times$} \\
64k & 24.633 & 23.981 & 14.877 & 1.66$\times$ & 7.272 & 5.309 & \textbf{1.37$\times$} \\
128k & 98.071 & 95.983 & 58.402 & 1.68$\times$ & 24.430 & 16.350 & \textbf{1.49$\times$} \\
256k & 391.466 & 384.460 & 229.349 & 1.71$\times$ & 87.756 & 54.412 & \textbf{1.61$\times$} \\
512k & 1579.215 & 1552.739 & 919.782 & 1.72$\times$ & 334.499 & 200.375 & \textbf{1.67$\times$} \\
\bottomrule
\end{tabular}
\end{table}

%% file: tables/tabicl_fp8.tex
\begin{table}[t]
\centering
\small
\caption{%
  FP8 ICL attention in \textbf{TabICLv2} against stock TabICLv2. The baseline scores an Elo of 1568.9. To estimate the accuracy impact of quantization from 16 bit to FP8, we do not employ our previous gating of $N>8192$. However, TabICLv2 already runs small datasets ($N<1024$) in FP32. Since our goal is to understand the impact of quantization from 16 bits to FP8, and to not confound our insights with this design, we only replace the attention in the existing 16 bit paths.
  Columns are as in Table~\ref{tab:appendix-fp8}.
}
\label{tab:tabicl-fp8}
\begin{tabular}{llrrrrr}
\toprule
& seed & $n$ & win & $\Delta$err (\%) & $\Delta$Elo & sign $p$ \\
\midrule
\multicolumn{7}{l}{\textit{TabArena, all splits (51 datasets)}} \\
fp8, single ICL call (ungated) & 0 & 46 & 0.43 & $+0.009$ & $-1.8$ & 0.231 \\
\bottomrule
\end{tabular}
\end{table}

%% file: tables/tabicl_speed.tex
\begin{table}[t]
\centering
\caption{%
  End to end speed of FP8 ICL attention in \textbf{TabICLv2} on one NVIDIA RTX PRO 6000 Blackwell Server Edition. Same setting as in \Cref{tab:speed}, however, for TabICLv2 we did not implement the actual gate at row count of 8k.%
}
\label{tab:tabicl-speed}
\begin{tabular}{rrrr}
\toprule
$N$ & baseline (s) & FP8 (s) & speedup   \\
\midrule
1k & 0.062 & 0.070 & 0.89$\times$   \\
2k & 0.093 & 0.097 & 0.96$\times$   \\
4k & 0.155 & 0.156 & 0.99$\times$   \\
\midrule
8k & 0.296 & 0.292 & \textbf{1.01$\times$}   \\
16k & 0.635 & 0.604 & \textbf{1.05$\times$}   \\
32k & 1.492 & 1.334 & \textbf{1.12$\times$}   \\
64k & 4.118 & 3.312 & \textbf{1.24$\times$}   \\
128k & 13.071 & 9.517 & \textbf{1.37$\times$}   \\
256k & 45.342 & 30.618 & \textbf{1.48$\times$}   \\
512k & 168.044 & 107.734 & \textbf{1.56$\times$}   \\
\bottomrule
\end{tabular}
\end{table}

%% file: tables/three_way_d128.tex
\begin{table}[t]
\centering
\caption{%
  A single attention call at head dimension $128$, on one NVIDIA RTX PRO 6000 Blackwell Server Edition.
}
\label{tab:three-way-d128}
\begin{tabular}{rrrrrr}
\toprule
& \multicolumn{3}{c}{time (ms)} & \multicolumn{2}{c}{FP8 speedup} \\
\cmidrule(lr){2-4}\cmidrule(lr){5-6}
$N$ & \texttt{torch} & ours 16-bit & ours FP8 & vs \texttt{torch} & vs 16-bit \\
\midrule
1k & 0.086 & 0.064 & 0.249 & 0.34$\times$ & 0.26$\times$ \\
2k & 0.146 & 0.148 & 0.285 & 0.51$\times$ & 0.52$\times$ \\
4k & 0.369 & 0.367 & 0.438 & 0.84$\times$ & 0.84$\times$ \\
\midrule
8k & 0.990 & 0.953 & 0.847 & \textbf{1.17$\times$} & 1.13$\times$ \\
16k & 3.544 & 3.426 & 2.337 & \textbf{1.52$\times$} & 1.47$\times$ \\
32k & 12.645 & 12.426 & 7.747 & \textbf{1.63$\times$} & 1.60$\times$ \\
64k & 50.588 & 50.063 & 29.792 & \textbf{1.70$\times$} & 1.68$\times$ \\
128k & 201.415 & 200.501 & 114.395 & \textbf{1.76$\times$} & 1.75$\times$ \\
256k & 811.438 & 811.546 & 450.350 & \textbf{1.80$\times$} & 1.80$\times$ \\
512k & 3273.265 & 3279.969 & 1800.439 & \textbf{1.82$\times$} & 1.82$\times$ \\
\bottomrule
\end{tabular}
\end{table}

%% file: tables/three_way_l4.tex
\begin{table}[t]
\centering
\caption{%
  A single attention call of our Triton kernel at head dimension $64$, on one NVIDIA L4.
  \texttt{torch} is the SDPA baseline path (FlashAttention-2).
}
\label{tab:three-way-l4-d64}
\begin{tabular}{rrrrrr}
\toprule
& \multicolumn{3}{c}{time (ms)} & \multicolumn{2}{c}{FP8 speedup} \\
\cmidrule(lr){2-4}\cmidrule(lr){5-6}
$N$ & \texttt{torch} & ours 16-bit & ours FP8 & vs \texttt{torch} & vs 16-bit \\
\midrule
1k & 0.251 & 0.165 & 0.859 & 0.29$\times$ & 0.19$\times$ \\
2k & 0.323 & 0.266 & 0.921 & 0.35$\times$ & 0.29$\times$ \\
4k & 0.676 & 0.618 & 1.144 & 0.59$\times$ & 0.54$\times$ \\
\midrule
8k & 2.273 & 2.220 & 2.051 & \textbf{1.11$\times$} & 1.08$\times$ \\
16k & 9.659 & 8.608 & 6.668 & \textbf{1.45$\times$} & 1.29$\times$ \\
32k & 38.241 & 35.354 & 24.784 & \textbf{1.54$\times$} & 1.43$\times$ \\
64k & 151.911 & 138.453 & 93.703 & \textbf{1.62$\times$} & 1.48$\times$ \\
128k & 604.884 & 555.437 & 363.973 & \textbf{1.66$\times$} & 1.53$\times$ \\
256k & 2613.455 & 2300.499 & 1443.517 & \textbf{1.81$\times$} & 1.59$\times$ \\
512k & 11799.523 & 10079.251 & 6254.635 & \textbf{1.89$\times$} & 1.61$\times$ \\
\bottomrule
\end{tabular}
\end{table}

%% file: references.bib
@misc{luo2026memoryefficienttabularfoundation,
      title={Memory Efficient Tabular Foundation Models}, 
      author={Shuting Luo and Monika Mikhail Kanaan and Cameron Gordon and Anna Leontjeva and Simon Lucey},
      year={2026},
      eprint={2607.27546},
      archivePrefix={arXiv},
      primaryClass={cs.LG},
      url={https://arxiv.org/abs/2607.27546}, 
}

@article{grinsztajn2026tabpfn,
  title={Tabpfn-3: Technical report},
  author={Grinsztajn, L{\'e}o and Fl{\"o}ge, Klemens and Key, Oscar and Birkel, Felix and Jund, Philipp and Roof, Brendan and Manium, Mihir and Hoo, Shi Bin and B{\"u}hler, Magnus and Garg, Anurag and others},
  journal={arXiv preprint arXiv:2605.13986},
  year={2026}
}

@article{shah2024flashattention,
  title={Flashattention-3: Fast and accurate attention with asynchrony and low-precision},
  author={Shah, Jay and Bikshandi, Ganesh and Zhang, Ying and Thakkar, Vijay and Ramani, Pradeep and Dao, Tri},
  journal={Advances in Neural Information Processing Systems},
  volume={37},
  pages={68658--68685},
  year={2024}
}

@misc{zadouri2026flashattention4algorithmkernelpipelining,
      title={FlashAttention-4: Algorithm and Kernel Pipelining Co-Design for Asymmetric Hardware Scaling}, 
      author={Ted Zadouri and Markus Hoehnerbach and Jay Shah and Timmy Liu and Vijay Thakkar and Tri Dao},
      year={2026},
      eprint={2603.05451},
      archivePrefix={arXiv},
      primaryClass={cs.CL},
      url={https://arxiv.org/abs/2603.05451}, 
}

@inproceedings{
erickson2026tabarena,
title={TabArena: A Living Benchmark for Machine Learning on Tabular Data},
author={Nick Erickson and Lennart Purucker and Andrej Tschalzev and David Holzm{\"u}ller and Prateek Mutalik Desai and David Salinas and Frank Hutter},
booktitle={The Thirty-ninth Annual Conference on Neural Information Processing Systems Datasets and Benchmarks Track},
year={2026},
url={https://openreview.net/forum?id=jZqCqpCLdU}
}

@inproceedings{
kubler2026when,
title={When {LLM}s get significantly worse: A statistical approach to detect model degradations},
author={Jonas M. K{\"u}bler and Kailash Budhathoki and Matth{\"a}us Kleindessner and Xiong Zhou and Junming Yin and Ashish Khetan and George Karypis},
booktitle={The Fourteenth International Conference on Learning Representations},
year={2026},
url={https://openreview.net/forum?id=cM3gsqEI4K}
}

@inproceedings{kurtic-etal-2025-give,
    title = "``Give Me {BF}16 or Give Me Death''? Accuracy-Performance Trade-Offs in {LLM} Quantization",
    author = "Kurtic, Eldar  and
      Marques, Alexandre Noll  and
      Pandit, Shubhra  and
      Kurtz, Mark  and
      Alistarh, Dan",
    booktitle = "Proceedings of the 63rd Annual Meeting of the Association for Computational Linguistics ",
    month = jul,
    year = "2025",
    address = "Vienna, Austria",
    publisher = "Association for Computational Linguistics",}

@article{vaswani2017attention,
  title={Attention is all you need},
  author={Vaswani, Ashish and Shazeer, Noam and Parmar, Niki and Uszkoreit, Jakob and Jones, Llion and Gomez, Aidan N and Kaiser, {\L}ukasz and Polosukhin, Illia},
  journal={Advances in neural information processing systems},
  year={2017}
}

@article{dao2022flashattention,
  title={Flashattention: Fast and memory-efficient exact attention with io-awareness},
  author={Dao, Tri and Fu, Dan and Ermon, Stefano and Rudra, Atri and R{\'e}, Christopher},
  journal={Advances in Neural Information Processing Systems},
  year={2022}
}

@article{dao2023flashattention,
  title={Flashattention-2: Faster attention with better parallelism and work partitioning},
  author={Dao, Tri},
  journal={arXiv:2307.08691},
  year={2023}
}

@inproceedings{qu2025tabicl,
  title={Tab{ICL}: {A} Tabular Foundation Model for In-Context Learning on Large Data},
  author={Qu, Jingang and Holzm{\"u}ller, David and Varoquaux, Ga{\"e}l and Le Morvan, Marine},
  booktitle={International Conference on Machine Learning},
  year={2025}
}

@inproceedings{
zhang2025sageattention,
title={SageAttention: Accurate 8-Bit Attention for Plug-and-play Inference Acceleration},
author={Jintao Zhang and Jia wei and Pengle Zhang and Jun Zhu and Jianfei Chen},
booktitle={The Thirteenth International Conference on Learning Representations},
year={2025},
url={https://openreview.net/forum?id=OL44KtasKc}
}

@inproceedings{
zhang2025sageattention2,
title={SageAttention2: Efficient Attention with Smoothing Q and Per-thread Quantization},
author={Jintao Zhang and Haofeng Huang and Pengle Zhang and Jia wei and Jun Zhu and Jianfei Chen},
booktitle={First Workshop on Scalable Optimization for Efficient and Adaptive Foundation Models},
year={2025},
url={https://openreview.net/forum?id=z1ph7lcmpv}
}

@article{qu2026tabiclv2,
  title={{TabICLv2}: {A} better, faster, scalable, and open tabular foundation model},
  author={Qu, Jingang and Holzm{\"u}ller, David and Varoquaux, Ga{\"e}l and Le Morvan, Marine},
  booktitle={International Conference on Machine Learning},
  year={2026}
}

@article{micikevicius2022fp8,
  title={Fp8 formats for deep learning},
  author={Micikevicius, Paulius and Stosic, Dusan and Burgess, Neil and Cornea, Marius and Dubey, Pradeep and Grisenthwaite, Richard and Ha, Sangwon and Heinecke, Alexander and Judd, Patrick and Kamalu, John and others},
  journal={arXiv:2209.05433},
  year={2022}
}

@article{frantar:2023:gptq,
    title={GPTQ: Accurate Post-Training Quantization for Generative Pre-trained Transformers}, 
      author={Elias Frantar and Saleh Ashkboos and Torsten Hoefler and Dan Alistarh},
    journal ={ICLR},
    year = {2023}
}

@article{purucker2026beyond,
  title={Beyond IID: How General Are Tabular Foundation Models, Really?},
  author={Purucker, Lennart and Tschalzev, Andrej and Erickson, Nick and Blayer, Gioia and Holzm{\"u}ller, David and Arazi, Alan and Pfefferle, Alexander and Tajjar, Mustafa and Varoquaux, Ga{\"e}l and Hutter, Frank and others},
  journal={arXiv preprint arXiv:2606.30410},
  year={2026}
}

@article{hollmann2025accurate,
  title={Accurate predictions on small data with a tabular foundation model},
  author={Hollmann, Noah and M{\"u}ller, Samuel and Purucker, Lennart and Krishnakumar, Arjun and K{\"o}rfer, Max and Hoo, Shi Bin and Schirrmeister, Robin Tibor and Hutter, Frank},
  journal={Nature},
  volume={637},
  number={8045},
  pages={319--326},
  year={2025},
  publisher={Nature Publishing Group UK London}
}

@inproceedings{parkInference2024,
author = {Park, Youngsuk and Budhathoki, Kailash and Chen, Liangfu and K\"{u}bler, Jonas M. and Huang, Jiaji and Kleindessner, Matth\"{a}us and Huan, Jun and Cevher, Volkan and Wang, Yida and Karypis, George},
title = {Inference Optimization of Foundation Models on AI Accelerators},
year = {2024},
booktitle = {Proceedings of the 30th ACM SIGKDD Conference on Knowledge Discovery and Data Mining},
}
